\documentclass[conference]{IEEEtran}
\IEEEoverridecommandlockouts
\usepackage{cite}
\usepackage{amsmath,amssymb,amsfonts}
\usepackage{algorithmic}
\usepackage{graphicx}
\usepackage{textcomp}
\usepackage{xcolor}
\def\BibTeX{{\rm B\kern-.05em{\sc i\kern-.025em b}\kern-.08em
    T\kern-.1667em\lower.7ex\hbox{E}\kern-.125emX}}

\title{Rapid Co-design of Task-Specialized Whegged Robots for Ad-Hoc Needs\\ }

\author{Varun Madabushi, Katie M. Popek, Craig Knuth, Galen Mullins, Brian A. Bittner
\thanks{Johns Hopkins University Applied Physics Laboratory, Laurel, MD, USA. {\tt\small varun.madabushi@jhuapl.edu}}}

\begin{document}
\maketitle

\begin{abstract}
In this work, we investigate the use of co-design methods to iterate upon robot designs in the field, performing time sensitive, ad-hoc tasks. 
Our method optimizes the morphology and wheg trajectory for a MiniRHex robot, producing 3D printable structures and leg trajectory parameters.
Tested in four terrains, we show that robots optimized in simulation exhibit strong sim-to-real transfer and are nearly twice as efficient as the nominal platform when tested in hardware.
\end{abstract}

\section{Introduction}
The era of industrial automation yielded the world’s first productive robots. 
These task-optimized robots were designed for pre-conceived, repetitive functions. 
Only recently have we seen robots leave the factory, mastering unstructured terrains.
These robots are task-generalists. 
However, they are often rendered unable to perform their mission in certain challenging environments. 
In this work, we hypothesize that these challenges can be mastered through minor changes to the morphology and control law of the \textit{task-generalist}, yielding a \textit{task-specialist} platform which can subsequently be fielded on the previously inaccessible task set.


To address this need, we look to the co-design methods space, which integrates morphology and control refinements into a single process that accounts for the holistic impacts of any design change.
Through co-design, an optimization framework can analyze changes in design parameters and offer suggestions to tune a design to a particular task.

Our main goal is to prove by demonstration (with n=4) that a robotic co-design framework can yield successful designs while achieving sample efficiency to enable in-field platform construction.
We share our philosophy towards designing a co-design architecture capable of achieving this goal, by selecting design parameters that (a) are compact for sample efficiency, (b) ensure satisfaction of basic task constraints (e.g. capable of standing stably), (c) are readily adjustable on a real robot, (d) maintain simulation fidelity to reality, and (e) express a rich variety of designs.

We implement this approach to co-design on a RHex \cite{saranli2001Rhex} robot, using simulations to optimize its morphology and control to maximize efficiency on a set of terrains.
We then fabricate the algorithmically-generated robots and evaluate their sim-to-real performance on these terrains. 
In this work, we will provide a brief overview of related work in co-design research and discuss our implementation, designed to improve platform performance on ad-hoc terrain sets.
We will close with thoughts on this work's implications for building fieldable co-design capabilities and plans for future work.

\begin{figure}
\label{fig:designs}
    \includegraphics[width=.5\textwidth]{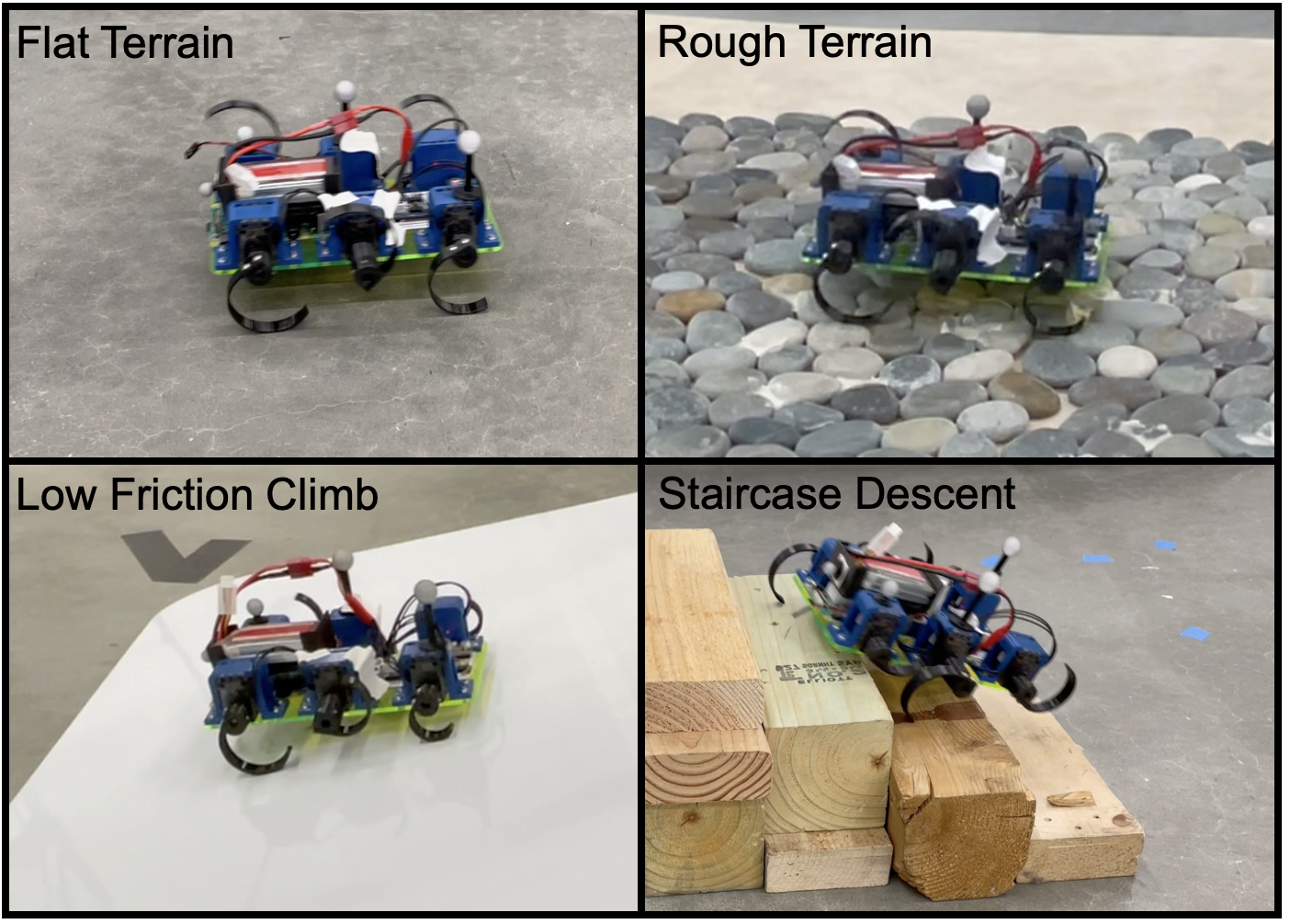}
    \caption{Four RHex co-designs were optimized in simulation over 100 trials in flat, rough, stairs, and ramp terrains through our approach, described in Section \ref{sec:method}. Each design outperforms the nominal gait in efficiency by 1.6x, 2x, 2.2x, and 5.1x respectively on hardware.}
\end{figure}

\section{Related Work}

Co-design frameworks integrate consideration of design and control variations. 
This problem is too large to solve in an exhaustive sense, so the methods covered here seek exploitative optimization techniques \cite{evans2022stochastic,rosendo2017trade} or reductions in problem representation \cite{carlone2019robot,zhao2020robogrammar}.
Several methods lump all design and control parameters into one large design space, representing the problem as a PDE or a Bayesian Optimization. 
This can result in quality high dimensional gradients that find competent platforms with sample efficiency. 
Some methods move toward assessing a discrete set of components \cite{carlone2019robot}, and in some cases graph grammars \cite{zhao2020robogrammar} for adding actuators and links to overall platform design.
In juxtaposition to high-dimensional robot parameterizations, we seek subtle refinements on small parameter spaces such that meaningful tasks can be added to a robot's capability set on timescales amenable to scenarios requiring rapid redesign in the field.

In observation of robot co-designs which lean on mechanical design rather than sophistication in the control law to achieve novel task sets \cite{stuart2014compliant,parness2009microfabricated,coad2019vine}, we seek to structure a co-design optimization that can meaningfully explore a region of the morphology space that can both better enable the task set and remove the burden on the control law. 

\section{Problem Statement}
We ask the robot to achieve maximum efficiency or speed on any terrain set. 
For a robot position $x \in SE(3)$ we take the final position $x_f$ and transform it back to the body frame of the initial location $x_i^{-1}x_f$, taking the first element of this transformation corresponding to the net displacement along the body over a fixed time frame. 
We obtain the forward efficiency by taking the average forward velocity over the average power (power computed as $\sum_{i=1}^6 \tau_i(t) \Omega_i$ in simulation and $\sum_{i=1}^6 V_i(t) I_i$ in hardware) throughout a trial. 
We do not ask the robot to achieve complex goals such as trajectory tracking.
Studies suggest that turning maneuvers to support trajectory tracking are more trivial to discover than forward locomotion and can be done rapidly in-situ \cite{bittner2021optimizing}.

\section{Method}
\label{sec:method}
Our parametrization of robot co-design and the optimization technique constitute our approach to co-design of a MiniRHex \cite{barragan2018minirhex} robot for high efficiency walking on a flat ground, across rough terrain, down a staircase, and up a slippery slope.
The MiniRHex features six compliant whegs, each operated by a single Dynamixel XL-320 servo.
To walk, each wheg tracks a pre-defined trajectory, consisting of a slow phase and a fast phase (corresponding to the swing phase as in bipeds \cite{grizzle2009mabel}).
The six whegs are grouped into two sets of three which follow the same trajectory offset half a cycle from each other. 
This forms a tripod which supports the robot during each step.

We invoke this structure on our open loop controller to simplify the control parametrization, which need only take four parameters (1) gait period, (2) duration of slow phase, (3) slow phase initial angle, and (4) slow phase termination angle.
We express the space of robot morphology variation with a similarly small set of parameters, by requiring bilateral symmetry to encourage stable straight-line motion.
The resulting choice of parameters includes a wheg stance defined by (5) front wheg length, (6) back wheg length, as well as (7) the thickness of the whegs.
Variation in the front and rear wheg lengths can adjust the ride height and weight distribution of the robot, while the stiffness of each wheg can be tuned by adjusting its thickness.
The co-design algorithm optimizes forward efficiency as defined in the problem statement using a bayesian optimizer designed for sample efficiency \cite{balandat2020botorch}.
This algorithm searches the parameter space through repeated trials, where the robot walks along the terrain of interest for 10s.
We restrict the optimizer to report the best co-design within 100 trials, generating solutions on the order of minutes or hours.

\section{Results}

\subsection{Simulation}
The optimization was conducted on the four terrains mentioned in Section \ref{sec:method}.
In order to improve numerical stability in the simulator, all dimensions were multiplied by 10.

Table \ref{tab:sim_eff} summarizes the results of optimizing the efficiency and speed of the MiniRHex robot in simulation where $\gamma$ is the efficiency of the platform and $v$ is the velocity of the platform.
For comparison, the nominal platform has efficiency 0.016 m/J and speed 0.52 m/s on flat ground.

\begin{table}[ht]
    \centering
    \begin{tabular}{c|c|c|c|c}
         Terrain & EOP $\gamma$(m/J) & EOP $v$(m/s) & SOP $\gamma$(m/J) & SOP $v$(m/s) \\ \hline
         Flat & \textbf{0.0248} & 0.685 & 0.00829 & \textbf{2.963} \\
         Rough & 0.0202 & 1.078 &  0.00522 & 2.521 \\
         Stairs & 0.0208 & \textbf{1.138} & \textbf{0.01527} & 2.183 \\
         Ramp & 0.0089 & 0.069 & 0.00182 & 1.276\\
    \end{tabular}
    \caption{Performance of Efficiency Optimized Platform (EOP) and Speed Optimized Platform (SOP) in Simulation}
    \label{tab:sim_eff}
\end{table}


\vspace{-.75cm}
\subsection{Hardware}
Each set of wheg and gait parameters was tested on physical examples of the previously mentioned terrains.
We evaluate the parameter changes by measuring the energy efficiency in m/J.

The distance traveled during each trial is measured through a set of OptiTrack motion tracking cameras, while the current and voltage are measured through a shunt current sensor, which is placed in series with the power supply.

\begin{figure}[ht]
    \centering
    \includegraphics[width=.5\textwidth]{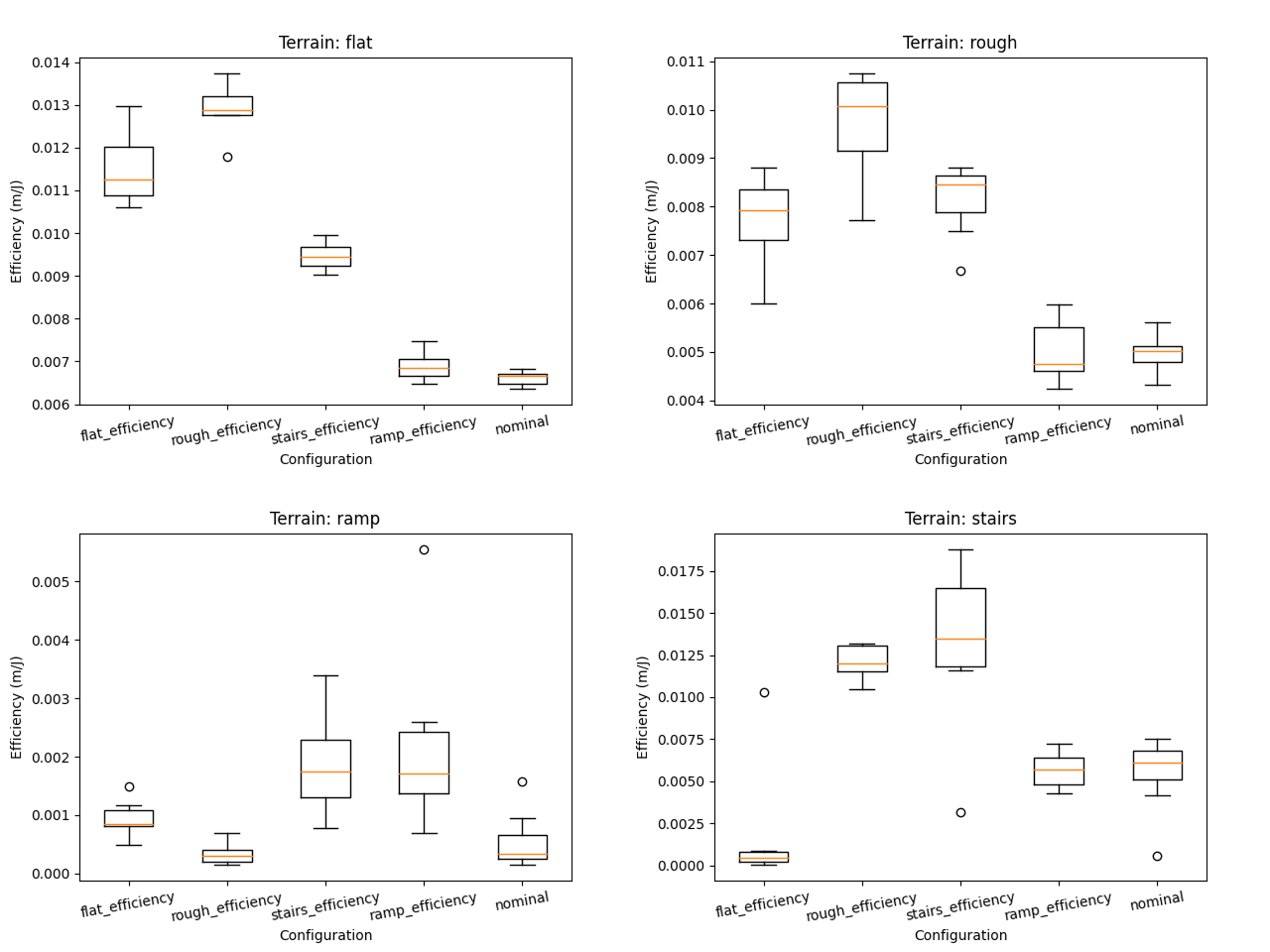}
    \caption{Efficiencies of each optimized platform on each terrain.
    All designs equalled or outperformed the nominal design, demonstrating co-design's capability to discover designs that would not be obvious to a human engineer.
    The optimization process resulted in a terrain-specialized platform which outperformed the nominal and other terrain platforms on its own environment.}
    \label{fig:cross_eval_plots}
\end{figure}

Each test was conducted 8 times for each of the efficiency-optimized robot designs, and the results are reported in Figure \ref{fig:cross_eval_plots}.
The performance of the Nominal MiniRHex (as-is, with no modifications) is also reported as a point of comparison.


On every terrain, all the optimized platforms performed equal to or better than the nominal platform in terms of efficiency, with the best platform demonstrating between 1.6x and 5x improvement over the nominal. 
Additionally, the best performer on each terrain was, with the exception of the flat-optimized robot, the robot optimized for that terrain, demonstrating that each co-designed robot developed specialized features to improve performance.

\section{Conclusions}
In this work, we claim to have proved by example that co-design is a potentially viable methodology for rapid refinement of designs \textit{in-situ}.
For each terrain of interest, the RHex robot refined both control and morphology parameters to improve efficiency.
On some terrains, such as the ramp, these innovations were required to achieve any locomotion.
The optimal efficiency gait for stair descent appears to engage in a controlled fall that dissipates little energy, whereas less efficient techniques tend to blunder down the stairs.
In this case, we hypothesize that the lack of feedback control guided the optimization toward leveraging passive stability in the design.
We find that the rough efficiency gait predictably sits on the robustness side of the efficiency-robustness tradeoff space by leveraging the use of larger whegs than the nominal and flat efficiency designs.
We found that our speed optimized gaits did not translate to reality, and expect this is due to higher impulse contact dynamics that are notoriously challenging to capture in multilegged platforms.
We were conversely encouraged by our sim-to-real transfer of efficiency designs, which leveraged lower impulse contact to conserve energy.
We also infer that the efficiency platform optimized for rough terrain may perform well on flat terrain due to its ability to reject disturbances in the dynamics.
Overall, we found these results to motivate further investigation of methods for rapid \textit{in-situ} development of robots to address critical ad-hoc needs.

\bibliographystyle{IEEEtran}
\bibliography{ref}

\begin{thebibliography}{10}
\providecommand{\url}[1]{#1}
\csname url@samestyle\endcsname
\providecommand{\newblock}{\relax}
\providecommand{\bibinfo}[2]{#2}
\providecommand{\BIBentrySTDinterwordspacing}{\spaceskip=0pt\relax}
\providecommand{\BIBentryALTinterwordstretchfactor}{4}
\providecommand{\BIBentryALTinterwordspacing}{\spaceskip=\fontdimen2\font plus
\BIBentryALTinterwordstretchfactor\fontdimen3\font minus \fontdimen4\font\relax}
\providecommand{\BIBforeignlanguage}[2]{{%
\expandafter\ifx\csname l@#1\endcsname\relax
\typeout{** WARNING: IEEEtran.bst: No hyphenation pattern has been}%
\typeout{** loaded for the language `#1'. Using the pattern for}%
\typeout{** the default language instead.}%
\else
\language=\csname l@#1\endcsname
\fi
#2}}
\providecommand{\BIBdecl}{\relax}
\BIBdecl

\bibitem{saranli2001Rhex}
U.~Saranli, M.~Buehler, and D.~E. Koditschek, ``Rhex: A simple and highly mobile hexapod robot,'' \emph{The International Journal of Robotics Research}, vol.~20, no.~7, pp. 616--631, 2001.

\bibitem{evans2022stochastic}
E.~N. Evans, A.~P. Kendall, and E.~A. Theodorou, ``Stochastic spatio-temporal optimization for control and co-design of systems in robotics and applied physics,'' \emph{Autonomous Robots}, vol.~46, no.~1, pp. 283--306, 2022.

\bibitem{rosendo2017trade}
A.~Rosendo, M.~Von~Atzigen, and F.~Iida, ``The trade-off between morphology and control in the co-optimized design of robots,'' \emph{PloS one}, vol.~12, no.~10, p. e0186107, 2017.

\bibitem{carlone2019robot}
L.~Carlone and C.~Pinciroli, ``Robot co-design: beyond the monotone case,'' in \emph{2019 International Conference on Robotics and Automation (ICRA)}.\hskip 1em plus 0.5em minus 0.4em\relax IEEE, 2019, pp. 3024--3030.

\bibitem{zhao2020robogrammar}
A.~Zhao, J.~Xu, M.~Konakovi{\'c}-Lukovi{\'c}, J.~Hughes, A.~Spielberg, D.~Rus, and W.~Matusik, ``Robogrammar: graph grammar for terrain-optimized robot design,'' \emph{ACM Transactions on Graphics (TOG)}, vol.~39, no.~6, pp. 1--16, 2020.

\bibitem{stuart2014compliant}
H.~S. Stuart, S.~Wang, B.~Gardineer, D.~L. Christensen, D.~M. Aukes, and M.~Cutkosky, ``A compliant underactuated hand with suction flow for underwater mobile manipulation,'' in \emph{2014 IEEE international conference on robotics and automation (ICRA)}.\hskip 1em plus 0.5em minus 0.4em\relax IEEE, 2014, pp. 6691--6697.

\bibitem{parness2009microfabricated}
A.~Parness, D.~Soto, N.~Esparza, N.~Gravish, M.~Wilkinson, K.~Autumn, and M.~Cutkosky, ``A microfabricated wedge-shaped adhesive array displaying gecko-like dynamic adhesion, directionality and long lifetime,'' \emph{Journal of the Royal Society Interface}, vol.~6, no.~41, pp. 1223--1232, 2009.

\bibitem{coad2019vine}
M.~M. Coad, L.~H. Blumenschein, S.~Cutler, J.~A.~R. Zepeda, N.~D. Naclerio, H.~El-Hussieny, U.~Mehmood, J.-H. Ryu, E.~W. Hawkes, and A.~M. Okamura, ``Vine robots,'' \emph{IEEE Robotics \& Automation Magazine}, vol.~27, no.~3, pp. 120--132, 2019.

\bibitem{bittner2021optimizing}
B.~Bittner and S.~Revzen, ``Optimizing gait libraries via a coverage metric,'' \emph{arXiv preprint arXiv:2107.08775}, 2021.

\bibitem{barragan2018minirhex}
M.~Barragan, N.~Flowers, and A.~M. Johnson, ``Minirhex: A small, open-source, fully programmable walking hexapod,'' in \emph{RSS Workshop}, vol.~8, 2018.

\bibitem{grizzle2009mabel}
J.~W. Grizzle, J.~Hurst, B.~Morris, H.-W. Park, and K.~Sreenath, ``Mabel, a new robotic bipedal walker and runner,'' in \emph{2009 American Control Conference}.\hskip 1em plus 0.5em minus 0.4em\relax IEEE, 2009, pp. 2030--2036.

\bibitem{balandat2020botorch}
M.~Balandat, B.~Karrer, D.~Jiang, S.~Daulton, B.~Letham, A.~G. Wilson, and E.~Bakshy, ``Botorch: A framework for efficient monte-carlo bayesian optimization,'' \emph{Advances in neural information processing systems}, vol.~33, pp. 21\,524--21\,538, 2020.

\end{thebibliography}

\appendices

\section{Supporting Simulation Fidelity of Wheg Compliance}
\label{sec:sim2real}
Having leg compliance parameters in simulation that relate with fidelity to what we can fabricate is crucial to successful design transfer to the operating environment.
However, our simulator Gazebo is incapable of simulating soft or deformable bodies.
Thus, in order to replicate and optimize over the compliant behavior of the wheg, we simulate an approximation of the curved wheg created from rigid links connected with spring-damper joints.

We compute the vertical displacement $\Delta y$ of the semicircular wheg tip under a contact force $F$ by applying Castigliano's theorem.
Castigliano's theorem states that the change in length of a beam equals the derivative of stored energy with respect to the external force.
\begin{align}
    U = \int_0^{L}\frac{M^2}{2EI} dl
    = \int_0^{\pi} \frac{(FR\sin{\theta})^2}{2EI} Rd\theta
    = \frac{\pi}{2}\frac{F^2R^3}{2EI}\\
    \Delta y = \frac{dU}{dF} = \frac{\pi F R^3}{2EI}  \\
    F = \frac{2EI}{\pi R^3} \Delta y = K\Delta y \label{Cont_Spring_Const}
\end{align}
$R$ is the wheg's radius, $M$ is the induced moment at each segment of the beam, $I$ is the area moment of inertia, and $E$ is the Young's modulus of the wheg material.

The result in equation \ref{Cont_Spring_Const} follows the structure of Hooke's law, demonstrating that the relationship between contact force $F$ and deflection $\Delta y$ can be modeled as a linear spring law with effective spring constant $K = \frac{2EI}{\pi R^3}$.

We then convert the effective spring constant of the wheg $K$ into a spring constant for each of the simulated wheg's rigid joints $K_T$, by treating it as a kinematic chain.
\begin{align}
\tau = J^{T}F = K_T \Delta \theta \\
K_T \Delta \theta = J^TF  
= J^TK
\begin{bmatrix}
    0\\
    \Delta y
\end{bmatrix} = KJ_{2}^T J_{2}\Delta \theta 
\end{align}
where $J$ is the Jacobian of the kinematic chain of the wheg link segments and $K_T$ is the set of spring constants. 
The composition of rigid links and spring joints with stiffness $K_T$ results in a structure which approximates the stiffness of the true wheg.
This approximation allows us to take a printable design variable (wheg thickness and length) and directly approximate what we expect the compliance to be in a rigid body simulator. 

\section{Cross Validation of Optimized co-designs}

Each platform is optimized as described in Section \ref{sec:method} on a specific terrain. Here, we take each design and test it in hardware on each environment, reporting the overall results in efficiency.

\end{document}